\documentclass[conference]{IEEEtran}
\IEEEoverridecommandlockouts
\usepackage{cite}
\usepackage{amsmath,amssymb,amsfonts}
\usepackage{algorithmic}
\usepackage{marginnote}
\usepackage{graphicx}
\usepackage{textcomp}
\usepackage{xcolor}
\usepackage{fancyhdr}
\usepackage{hyperref} 
\def\BibTeX{{\rm B\kern-.05em{\sc i\kern-.025em b}\kern-.08em
    T\kern-.1667em\lower.7ex\hbox{E}\kern-.125emX}}
\begin{document}

\title{Explainable LightGBM Approach for Predicting Myocardial Infarction Mortality\\
}

\author{\IEEEauthorblockN{1\textsuperscript{st} Ana Letícia Garcez Vicente}
\IEEEauthorblockA{\textit{Computer Science Department} \\
\textit{University of São Paulo}\\
São Carlos, Brazil \\
analeticiagarcez@usp.br}
\and
\IEEEauthorblockN{2\textsuperscript{nd} Roseval Donisete Malaquias Junior}
\IEEEauthorblockA{\textit{Computer Science Department} \\
\textit{University of São Paulo}\\
São Carlos, Brazil \\
roseval@usp.br}
\and
\IEEEauthorblockN{3\textsuperscript{rd} Roseli A. F. Romero}
\IEEEauthorblockA{\textit{Computer Science Department} \\
\textit{University of São Paulo}\\
São Carlos, Brazil \\
rafrance@icmc.usp.br }

}

\maketitle

\begin{abstract}
Myocardial Infarction is a main cause of mortality globally, and accurate risk prediction is crucial for improving patient outcomes. Machine Learning techniques have shown promise in identifying high-risk patients and predicting outcomes.  However, patient data often contain vast amounts of information and missing values, posing challenges for feature selection and imputation methods. In this article, we investigate the impact of the data preprocessing task and compare  three ensembles boosted tree methods to predict the risk of mortality in patients with myocardial infarction. Further, we use the Tree Shapley Additive Explanations method to identify relationships among all the features for the performed predictions, leveraging the entirety of the available data in the analysis. Notably, our approach achieved a superior performance when compared to other existing machine learning approaches, with an F1-score of 91,2\% and an accuracy of 91,8\% for LightGBM without data preprocessing.
\end{abstract}


\section{Introduction}

Since around the 1990s, Cardiovascular disease (CVD) has been the main cause of death and according to the World Health Organization, around 18 million people die from CVDs yearly \cite{owidcausesofdeath}. 
Among CVDs, Acute Myocardial Infarction (AMI), colloquially known as a heart attack, accounts for the largest percentage of deaths, suggesting it is a core issue in cardiovascular disease. Myocardial infarction (MI) occurs when blood flow decreases or stops in the coronary artery of the heart, causing possibly irreparable damage to the heart muscle. In its first year, AMI is associated with a high mortality rate and its incidence remains high in all countries. In the United States, more than one million people suffer from MI every year and about 20\% to 30\% of them die from AMI before arriving at the hospital. Almost half of MI patients suffer severe complications that may lead to death, so MI cases cannot be left unrecognized and untreated. 
Several studies have proposed methods of MI early identification as in \cite{w1} and some are focused on finding biomarkers as \cite{w2}. Also, there are studies that are looking specifically at the complications, including mortality of MI, such as in \cite{w3}, \cite{w4}, and \cite{w5}. Most works for predicting mortality from MI focus on exploring the best data preprocessing techniques for model adjustment. To the best of our knowledge, no studies were found that explore the use of models based on ensemble boosted trees, such as XGBoost \cite{chen2015xgboost}, LightGBM \cite{ke2017lightgbm}, and CatBoost \cite{prokhorenkova2018catboost}. 

Thus, this work aims to develop Machine Learning (ML) models for predicting mortality in patients admitted to the hospital with MI. We mainly utilize three ensembles boosted tree methods, namely XGBoost, LightGBM, and CatBoost, incorporating all available data, including information collected at admission and at 24, 48, and 72 hours after the onset of infarction. We evaluate if these methods eliminate the need for imputation, feature selection, and target feature balancing, simplifying the application of the models and improving its training time. We propose that the use of simple algorithms in the feature selection, imputation, and target feature balancing stage may limit the performance of these methods, as they themselves perform feature selection and handle missing values and imbalanced datasets. This also hinders the interpretability analysis of the models, as they may remove important features from the analysis.

In order to validate our hypothesis that the performance of these models is limited by the use of straightforward algorithms for attribute selection, imputation, and target feature balancing, we carried out an ablation study. Within this study, we deliberately disabled specific components of the preprocessing pipeline for the ensemble boosted tree models, allowing us to conduct a comparative analysis of the performance obtained both with and without preprocessing. 

By making all data accessible to the predictive model, it becomes feasible to employ interpretative tree-based techniques like Tree SHAP (SHapley Additive exPlanations) \cite{shap} to examine the impact of all attributes on classification. Thus, our work emphasizes explainability, which holds significant importance in the medical field \cite{vellido2020importance}. We employed the Tree SHAP method for the best model produced to establish connections among all the features in the dataset during the prediction process, utilizing the complete set of available data in the analysis. This emphasis on explainability ensures that decisions made using our models are transparent and comprehensible for healthcare professionals, and, consequently, facilitate and increase the medical adoption of these models. 

This article is organized as follows. First, in section \ref{relatedwork} we present previous related work. In section \ref{methods} is explored the dataset, preprocessing steps applied, and classification methods. Section \ref{experiments} displays the experiments made and the results obtained as well the interpretability of the results obtained. Finally, the conclusion and present potential directions for future research are presented in section \ref{conclusion}. 

\section{Related Work}\label{relatedwork}

ML techniques have become increasingly prevalent in the field of medicine, including myocardial infarction. Researchers have sought to address the crucial task of early identification of infarction cases. For instance, in the study \cite{w1}, the authors aimed to utilize ML for the early prediction of acute myocardial infarction and patient mortality among individuals experiencing chest pain. To achieve this, they employed a combination of electrocardiogram (ECG) exams and blood tests, leveraging the power of Convolutional Neural Networks (CNNs). Remarkably, their approach achieved an accuracy rate of 60\% in correctly identifying patients.

Furthermore, there have been notable endeavors to predict mortality associated with myocardial infarction. In the studies \cite{w4} and \cite{w5}, the primary objective was to predict mortality using various ML approaches. Remarkably, these studies achieved accuracies of 80\% and 86.74\% respectively. 

Other studies have focused on predicting complications that may arise after the occurrence of myocardial infarction. In the study \cite{w3}, the authors aimed to predict specific complications following myocardial infarction. They employed a Multi-Layer Neural Model, coupled with explainable tools, to forecast potential infarction-related complications.

Numerous instances of medical data are disseminated in heterogeneous tabular formats, a characteristic illustrated by myocardial examinations, as referenced in \cite{w1}, \cite{w4}, \cite{w5}, and \cite{w3}. Nevertheless, the inherent heterogeneity poses a considerable obstacle to the practical application of deep learning algorithms, as these algorithms commonly necessitate preprocessing, a step susceptible to introducing bias into the data, as elucidated by \cite{borisov2022deep}. Vadim's research, in particular, conducted a comparative evaluation of deep learning algorithms against tree boosting-based techniques, such as XGBoost, LightGBM, and CatBoost, revealing the latter to exhibit superior performance characteristics. 

In addition to the preprocessing challenges, the medical domain grapples with the integration of "black box" predictive methods. In this context, interpretability plays a crucial role in securing acceptance from both patients and medical experts \cite{vellido2020importance}. To address the need for transparency in these "black box" methods, the Tree SHAP technique has emerged as a promising solution \cite{shap}. In a specific case \cite{9288053}, a binary prediction model employing XGBoost was developed for cardiovascular disease in patients with Type 2 Diabetes Mellitus. The application of the Tree SHAP method was instrumental in enhancing the model's transparency and instilling confidence in its suitability for the medical field. This method quantifies attribute influence in predictions, facilitating a more understandable and trustworthy model in the complex landscape of medical data analysis.

\section{Methods}\label{methods}

\subsection{Dataset}
The Myocardial Infarction Complications dataset \cite{dataset-paper1} \cite{dataset-paper2} was obtained from the Krasnoyarsk Interdistrict Clinical Hospital in Russia during the 1990s by the University of Leicester. However, it was only made publicly available in 2020. This dataset comprises data collected from patients who were admitted to the hospital with myocardial infarction. In addition to patient-specific information, it includes data points related to the following time intervals: i) time of admission, ii) end of the first day (24 hours), iii) end of the second day (48 hours), and iv) end of the third day (72 hours). These data points enable the tracking of disease progression in individual patients, considering that the course of the disease can vary and lead to different outcomes and complications. 

The dataset consists of 1700 instances, each comprising 124 features or attributes. Among these attributes, 111 are input characteristics, including numerical, binary, and categorical ordinal data, while the remaining 12 serve as potential target labels. Approximately 7.6\% of the instances contain missing values. Regarding the target class of this work, which identifies the complications leading to mortality accounting for 15.94\% of the instances, and it encompasses categories such as Cardiogenic Shock, Pulmonary Edema, Myocardial Rupture, Progress of Congestive Heart Failure, Thromboembolism, Asystole, and Ventricular Fibrillation. The remaining 84.06\% correspond to individuals who survived.

\subsection{Preprocessing Task} \label{pipeline}

Initially, we performed data cleaning by removing features with more than 10\% missing values. We also removed the columns with single value dominance of more than 95\%, resulting in a dataset comprising 61 columns, excluding the ID and target class. To facilitate further analysis, we binarized the target class, combining various lethal complications into a single category, resulting in Deceased (271 instances) and Unknown (Alive) classes (1429 instances), which still results in an unbalanced dataset. Before proceeding with additional preprocessing steps to prevent data contamination, we divided the dataset into training and testing sets, allocating 80\% for training and 20\% for testing purposes. 

We applied Random Undersampling, adjusting the imbalance between the majority class (denoted as $N_{Maj}$) and the minority class (denoted as $N_{Min}$) by setting $N_{Min} = \alpha N_{Maj}$. We experimented with different values of $\alpha_{RUS}$, including 0.5, 0.8, and 1.0, to explore the impact of varying levels of undersampling on the predicting task. However, as highlighted in \cite{w5}, which investigates the optimal preprocessing techniques for the dataset in question, it was observed that Random Undersampling with a ratio of 0.5 yields superior performance for predictive models in this particular context. Consequently, this sampling algorithm with the specified ratio was selected as our dataset balance method. 

During the preprocessing task, the dataset was manipulated to convert all attributes into a numerical format, encompassing both binary and categorical ordinal data. Consequently, we utilized One Hot Encoding for categorical nominal columns and employed Label Encoding for categorical ordinal data. Furthermore, in order to ensure consistency and comparability among features, we applied a normalization technique, which scaled the values within a range of 0 to 1. 

Finally, we addressed the problem of missing data using two methods. For categorical data, we filled in the missing values with the mode, which is the most frequently occurring value in that attribute. For numerical data, we replaced the missing values with the median, calculated from the available values in the corresponding attribute.

Considering the extensive collection of over 61 attributes in the dataset (following the initial preprocessing steps), conducting a manual analysis of each individual feature would be a laborious task susceptible to human error. Consequently, achieving an optimal feature selection would require domain-specific knowledge. Therefore, it was crucial to employ a feature selection algorithm to mitigate these challenges, minimize computational complexity, and tackle the risk of overfitting. In this regard, we adopted a method to select the K most significant features that have the greatest linear correlation on the target class. This algorithm functions by identifying the K features with the highest scores, determined through the $Chi^2$ test conducted between each feature and the target class.

\subsection{Model Selection}

To determine the optimal model for mortality prediction, we conducted a thorough evaluation using a grid search with cross-validation, employing a 5-fold approach on the 80\% training data. We assessed multiple models, including Multi-Layer Perceptron (MLP), Support Vector Machine (SVM), Logistic Regression (LR), Decision Tree (DT), Random Forest (RF), as well as ensemble boosted trees algorithms such as XGBoost, LightGBM, and CatBoost. We systematically explored a range of model parameters through a comprehensive grid search process. Furthermore, we employed the SelectKBest algorithm to perform feature selection, considering feature counts of 15, 35, and 50. It is important to highlight that these hyperparameters for the preprocessing pipeline were fine-tuned alongside the hyperparameters of the prediction models during the grid search process.

Throughout the grid search, we assessed the models based on the F1 score metric. Due to the imbalanced nature of the data, our priority was to prioritize metrics such as recall and F1 score over accuracy. This approach aligns with our objective of minimizing false negatives, as our goal is to avoid misclassifying high-risk patients as low-risk in the medical context. By doing so, we ensure that individuals with critical conditions receive the necessary attention and appropriate medical interventions. Moreover, by taking into account the F1 score of the models, we ensure a balance between recall and precision. It is of utmost importance to consider this metric, as solely concentrating on recall can potentially mislead us in identifying a ``false'' optimal model. In such scenarios, the model might predict all examples as positive (overfitting), yielding a recall of 100\%. However, the precision would be low, rendering the model ineffective. 

In conclusion, each model with the most optimal hyperparameter settings underwent evaluation through a 10-fold cross-validation on the training set (80\%). The evaluation encompassed metrics such as accuracy, recall, precision, and F1-score. Subsequently, the best model was selected based on these assessed metrics and its practical relevance within the proposed real world scenario. The best model was then trained using the training set (80\%) and subsequently assessed using the test set (20\%). Lastly, a quantitative analysis was conducted on the metrics of the final model, aiming to determine the viability of applying the predictive model in a real-world context. More details can be seen on the following Github page: \url{https://github.com/analeticiagarcez/Myocardial-infarction-prediction}.

\section{Experiments and Results}\label{experiments}

Initially, a grid search was conducted to explore different combinations of parameters. The grid search yielded the optimal hyperparameters based on the F1 Score, and subsequently, each model was retrained with the respective parameters. For this phase, a 10-fold cross-validation approach was employed. The comprehensive outcomes of this process can be observed in Table 1. 

Given the practical implications of the predictive model in a real world setting, particularly in the domain of suggesting immediate hospitalization for AMI patients, it becomes imperative to minimize the occurrence of false negatives (i.e., individuals who require hospitalization but are incorrectly discharged). Consequently, achieving a high recall value becomes a desirable attribute for the predictive model's effectiveness. Nonetheless, it is worth emphasizing that a comprehensive analysis encompassing all relevant metrics is essential to comprehensively assess the performance and behavior of the models under consideration. 

Considering the calculated metrics and the practical application context of the model, the RF was designated as the superior trained model. It showcases the highest recall compared to all other evaluated models, alongside a commendable F1-Score value, signifying a harmonious balance between precision and recall metrics. The RF model exhibits the highest recall value, indicating a superior ability to minimize false negatives within the scope of the eight evaluated models. Given that false negatives have significant implications in the practical application of the predictive model, potentially leading to the improper discharge of patients requiring hospitalization, the RF model was selected as the final model for evaluation using the test set (20\%). The final hyperparameters for the RF model were: Criterion = entropy, Max features = None, Min samples leaf = 4, Min samples split = 2, and for the feature selection the best was k = 50. 

\begin{table}[h]
\caption{10-fold cross-validation results}
\begin{center}
\begin{tabular}{ccccc}
\hline
\multicolumn{5}{c}{Pipeline 1}  \\
\hline
\textbf{Model} & \textbf{wF1} & \textbf{wPrecision} & \textbf{wRecall} & \textbf{Accuracy} \\
\hline
RF          & 0.920   & 0.922   & 0.924   & 0.924 \\ \hline 
CatBoost    & 0.916   & 0.918   & 0.917   & 0.917 \\ \hline 
LightGBM    & 0.907   & 0.911   & 0.910   & 0.910 \\ \hline 
XGBoost     & 0.895   & 0.901   & 0.893   & 0.893 \\ \hline 
MLP         & 0.894   & 0.901   & 0.891   & 0.891 \\ \hline 
LR          & 0.890   & 0.896   & 0.888   & 0.888 \\ \hline 
SVM         & 0.888   & 0.913   & 0.907   & 0.907 \\ \hline 
DT          & 0.846   & 0.874   & 0.834   & 0.834 \\ \hline 
\hline
\multicolumn{5}{c}{Pipeline 2}  \\
\hline
\textbf{Model} & \textbf{wF1} & \textbf{wPrecision} & \textbf{wRecall} & \textbf{Accuracy} \\
\hline
CatBoost     & 0.917   & 0.918   & 0.917   & 0.917  \\ \hline 
LightGBM     & 0.922   & 0.924   & 0.926   & 0.926  \\ \hline 
XGBoost      & 0.918   & 0.918   & 0.918   & 0.918  \\ \hline
\end{tabular}
\label{tab1}
\end{center}
\end{table}

\subsection{Ablation Study}

As evident from the preceding discourse, the ensemble boosted tree models showcased exceptional performance in comparison to other ML models, with the exception of the Random Forest model. To delve deeper into the analysis of these approaches, we examined two distinct pipelines for the dataset and observed their respective outcomes. 

In Pipeline 1, we adhered to the methodology outlined in Section \ref{pipeline}, which encompassed executing specific procedures for data preprocessing and feature selection as described. In contrast, Pipeline 2 utilized the raw dataset without any alterations, employing the data exactly as it was received. Upon examining Table 1, we noted that the models trained with the raw data outperformed those trained with preprocessed data. This observation suggests that these methods obviate the need for imputation, feature selection, and target feature balancing. Conversely, the utilization of simple algorithms for feature selection, imputation, and target feature balancing may constrain the performance of these methods, as they inherently incorporate these functionalities. Nonetheless, drawing definitive conclusions necessitates conducting a statistical test to determine whether the metrics obtained between the models with and without preprocessing exhibit statistically significant differences.

Based on the obtained results, it became evident that all models utilizing Pipeline 2 showcased superior or comparable performance compared to the models employing the first pipeline. To ascertain the significance of these findings, a statistical paired t-test was conducted to compare the F1-score values across the models. Interestingly, it was observed that LightGBM, CatBoost, and XGBoost performed on par with the model trained without preprocessing, yielding p-values greater than 0.05 ($p = 0.703$ for LightGBM, $p = 0.632$ for XGBoost, and $p = 0.846$ for CatBoost). Furthermore, when comparing the best model with preprocessing (RF), to the best model without preprocessing (LightGBM), no statistically significant difference in performance was observed ($p > 0.05$ with $p = 0.986$) based on the results of the paired t-test.

This study was particularly intriguing due to the presence of a model that doesn't require preprocessing. This factor can significantly impact the robustness of the algorithm, as it exhibits greater resilience and adaptability to variations when tested with new input data. Preprocessing often involves assumptions about data distribution and imposes specific transformations, which can directly affect the model's performance in real scenarios if the new data undergoes changes. By eschewing preprocessing, the model benefits from enhanced flexibility and consistently reliable performance. It also improves the model's generalization capability, allowing it to effectively handle new and previously unseen examples, an invaluable trait in real-world applications. Moreover, the simplicity of the approach aids in model implementation as well as usability and facilitates algorithm interpretation.

\subsection{Final Evaluation}

Finally, LightGBM model without preprocessing and RF model were selected as the best models. While no statistically significant difference was found between them, it is evident that the LightGBM model exhibits superior metrics in the 10-fold cross-validation. Hence, both models were evaluated using the test dataset to simulate real-world application scenarios and assess their performance on unseen data. The obtained metrics in Table 2 closely resemble those achieved during the 10-fold cross-validation, suggesting that the models may consistently perform well in practical contexts. 

Based on the results obtained from the final predictive model, it is suitable for recommending hospitalization of AMI patients, supported by its high recall value of $0.903$ and F1-Score of $0.912$, which indicate the model's reliability.

\begin{table}[h]
\caption{Test results}
\begin{center}
\begin{tabular}{cccccc}
\hline
\textbf{Model}&\textbf{Pipeline} & \textbf{wF1} & \textbf{wPrecision} & \textbf{wRecall} & \textbf{Accuracy} \\
\hline
RF         & 1 & 0.900   & 0.899   & 0.903   & 0.903 \\ 
\hline
LightGBM   & 2 & 0.912   & 0.914   & 0.918   & 0.918 \\  
\hline
\end{tabular}
\label{tab1}
\end{center}
\end{table}

\subsection{Model Interpretation}

To gain insights into the model predictions and identify the key features influencing the final results, we utilized the SHAP (Shapley Additive Explanations) method on one of our best models, the LightGBM with Pipeline 2. Visualizing and exploring the data play a critical role in uncovering significant patterns within datasets, facilitating efficient knowledge discovery and analysis. Capitalizing on these advantages, we conducted a comprehensive visual exploration to examine the correlation between mortality and input characteristics.

The SHAP method is grounded in game theory and utilizes Shapley values. Shapley values represent the contribution of each player in a cooperative game. In a similar vein, the SHAP model measures the contribution of each feature to the final prediction made by a Machine Learning model. In other words, the SHAP values for a specific feature represent the prediction difference when the feature is or is not included. For this particular study, we opted for the Tree SHAP algorithm, which is specifically designed for tree-based algorithms.

By considering the target class as the outcome, 
Figure 1 
provide an overview of the impact of each feature on the predicted outcome, as estimated by the Shapley values. Figure 1 represents the data using a beeswarm plot, where each dot represents the Shapley value for a single observation. The color of the dot corresponds to the value of the respective feature. The features are arranged by relevance, allowing us to conclude that the most significant factors influencing the prediction are systolic blood pressure, the time elapsed from the beginning of the attack to hospital admission, and the relapse of pain on the third day.

Specific features such as the relapse of pain on the third day, age, white blood cell count, and the presence of infarction in the left ventricular area are positively associated with the risk of mortality. Conversely, other features such as the elapsed time and the use of opioids in the emergency department are negatively related to the risk. It's interesting to note from the plot that the shorter the time taken by a person to reach the hospital after the attack, the lower their risk of mortality. This suggests that if a person successfully makes it through the initial critical period, their chances of survival increase.

\begin{figure}[]
\centerline{\includegraphics[width=0.5\textwidth]{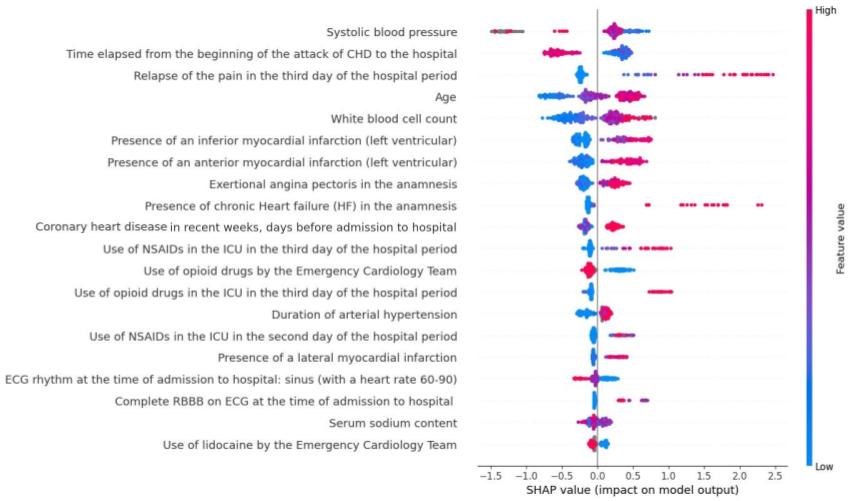}}
\caption{Shapley Values showing the influence exerted by the most influential features in predicting mortality}
\label{fig}
\end{figure}

\section{Conclusion}\label{conclusion}

This article conducted an extensive investigation into various Machine Learning methods, with a specific focus on boosted tree techniques. Our investigation unveiled the remarkable effectiveness of this approach and achieved higher values compared to those demonstrated in previous studies. Additionally, our hypothesis postulated that the utilization of simple algorithms in the feature selection, imputation, and target feature balancing stage may constrain the performance of these methods, as they inherently deal with feature selection,  missing values, and imbalanced datasets. Thus,  an ablation study was also conducted to assess the performance of boosted tree models using two distinct pipelines. The first pipeline involved preprocessing, while the second pipeline omitted any preprocessing steps. Intriguingly, we observed that the models without preprocessing exhibited the best performances. Among them, LightGBM method emerged as the top performer on the test set, attaining an impressive F1 score of $91.2\%$ and an accuracy of $91.8\%$. Finally, from the data analysis done thanks to use of the LighGBM,   the most significant factors influencing the prediction are systolic blood pressure, the time elapsed from the beginning of the attack to hospital admission, and the relapse of pain on the third day.

As future works, we intend to get more data containing updated information and validate our hypothesis considering other machine learning techniques.

\section{Acknowledgment}\label{acknowledgment}

This work was partially financed by the Coordenação de Aperfeiçoamento de Pessoal de Nível Superior - Brasil (CAPES), and by Fundação de Amparo à Pesquisa do Estado de São Paulo (FAPESP) \#
2023/06737-7 and  INCT
(CAPES \#88887.136349/2017-00, CNPQ \#465755/2014-3).

\bibliographystyle{ieeetr}
\bibliography{bibli}

\end{document}